\documentclass{svproc}
\usepackage{emptypage}  
\usepackage{cite}
\usepackage{amsmath,amssymb,amsfonts}
\usepackage{algorithmic}
\usepackage{graphicx}

\usepackage{textcomp}
\usepackage{subcaption}
\usepackage{xcolor}
\usepackage{array}
\usepackage{enumitem}

\usepackage[section]{placeins}
\usepackage{cite}
\usepackage{graphicx}
\usepackage{url}
\usepackage{amsmath,amssymb}
\usepackage{float}

\usepackage{booktabs}
\usepackage{caption}
\usepackage{graphicx}
\usepackage{caption}
\usepackage{tikz}
\usepackage{fancyhdr}
\usepackage[utf8]{inputenc}
\usepackage{geometry}
\usepackage{underscore}
\usepackage{booktabs}

\usepackage{orcidlink}
\newcommand{\orcidID}[1]{\orcidlink{#1}}

\usepackage{hyperref}
\usetikzlibrary{calc,positioning, shapes.geometric, arrows.meta}

\begin{document}
\mainmatter

\title{Detection of Adversarial Attacks in Robotic Perception}

\author{
Ziad Sharawy\orcidID{0009-0008-3502-1510} \and
Mohammad Nakshbandi\orcidID{0009-0009-4870-0958} 
Sorin Mihai Grigorescu\orcidID{0000-0003-4763-5540}
}

\institute{
Department of Mechatronics and Robotics,\\
Faculty of Electrical Engineering and Computer Science,\\
Transylvania University of Brașov, Romania\\
\email{ahmed.sharawy@unitbv.ro,
mohammed.nakshbandi@unitbv.ro,
s.grigorescu@unitbv.ro}
}

\maketitle

 Ziad Sharawy, Mohammad-Maher Nakshbandi and Sorin Grigorescu are with the Robotics, Vision and Control Laboratory (RovisLab,
\url{https://www.rovislab.com}), Transilvania University of Brașov, Romania. 
The GitHub code is available at \url{https://github.com/RovisLab/CyberAI_Ziad}.
\begin{abstract}
Deep Neural Networks (DNNs) achieve strong performance in semantic segmentation for robotic perception but remain vulnerable to adversarial attacks, threatening safety-critical applications. While robustness has been studied for image classification, semantic segmentation in robotic contexts requires specialized architectures and detection strategies.\par

We propose a framework for detecting adversarial attacks using pre-trained ResNet-18 and ResNet-50 models. Our method leverages advanced feature extraction and statistical metrics to distinguish clean from adversarial inputs. Experiments demonstrate its effectiveness across various attacks, offering insights into model robustness. Additionally, we compare network architectures to identify factors that enhance resilience. This work supports the development of secure autonomous systems by providing practical detection tools and guidance for selecting robust segmentation models.

\end{abstract}

\section{Introduction}
\label{sec:introduction}
The field of computer vision has transformed over the past decade, driven by deep learning advances \cite{lecun2015deep, goodfellow2016deep}. Semantic pixel-wise segmentation, which assigns a class label to every pixel \cite{long2015fully}, is critical for applications such as autonomous driving \cite{chen2018deeplab}, medical imaging \cite{ronneberger2015u}, urban planning, and robotics, where accurate visual understanding impacts decision-making. Traditional segmentation based on hand-crafted features often fails in complex scenarios \cite{hariharan2015hypercolumns}, whereas deep CNNs enable hierarchical feature extraction, significantly improving performance \cite{he2016deep, krizhevsky2012imagenet}.

In safety-critical systems like autonomous vehicles and robotics, robust perception must withstand adversarial attacks—carefully crafted perturbations designed to mislead models \cite{szegedy2013intriguing, goodfellow2014explaining}. Such attacks can cause misclassifications, potentially leading to failures and eroding trust \cite{papernot2016limitations, carlini2017towards}. Detecting adversarial inputs is therefore crucial for safety and reliability.

This work targets adversarial detection in robotic visual perception by extending pre-trained ResNet-18 and ResNet-50 \cite{he2016deep} for dense semantic feature extraction. By combining advanced feature extraction with novel detection strategies \cite{miller2017ada_detection, xu2017feature}, our framework distinguishes original from adversarial images, enhancing segmentation reliability and mitigating adversarial risks.

Overall, this contribution promotes secure and trustworthy autonomous systems, emphasizing proactive adversarial detection for safer AI deployment in critical applications.


\section{Related Work}
\label{sec:related_work}

Adversarial-example detection complements robustness-based defenses by identifying maliciously perturbed inputs rather than directly making models attack-resistant. Early statistical approaches detect adversarial inputs via distributional shifts in network activations \cite{grosse2017statistical_detection}, while anomaly-detection frameworks treat adversarial examples as outliers in feature space \cite{miller2017ada_detection}. Model-uncertainty-based methods use predictive entropy and mutual information to flag inputs in low-confidence regions \cite{smith2018uncertainty_detection}. Robust detection strategies combining input transformations and feature squeezing compare predictions across transformed inputs to detect inconsistencies caused by adversarial perturbations \cite{ma2019robust_detection}.

Graph-based methods model relationships in the network’s latent space. Latent neighborhood graphs of activations enable detection through disruptions in the manifold structure of benign examples \cite{abusnaina2021lng_detection}. Frequency-domain analyses exploit high-frequency perturbations, allowing reconstruction-based anomaly detection \cite{freqdomain2024_detection}. Similarly, universal detectors compare deep features across layers against reference clean examples \cite{mumcu2024detecting_adversarial}, while semantic-aware approaches, such as semantic graph matching, detect perturbations disrupting contextual object relationships \cite{semantic_graph_match2023_detection}. In text, embedding-level and syntactic consistency checks identify adversarial manipulations \cite{text_adv_detection2022}.

Despite these advances, many methods remain vulnerable to adaptive attacks \cite{carlini2017adversarial_not_detected}, motivating our work to improve reliability against both known and unseen attack types.

\section{Method}
\label{sec:metod}

\subsection{Problem definition}

This paper addresses separating original and adversarial images using pre-trained ResNet-18 and ResNet-50 for semantic segmentation. Our contribution is a specialized adversarial detection method for robotic perception, leveraging feature extraction tailored for segmentation tasks and evaluating multiple attack types to assess model resilience in safety-critical applications.

Semantic segmentation assigns a class label to each pixel. For input $I \in \mathbb{R}^{H \times W \times C}$, a deep network extracts low-level (edges, textures) and high-level (object shapes, context) features, producing a label map $L \in \mathbb{R}^{H \times W \times K}$, where $L_{i,j}$ gives the predicted class of $I_{i,j}$. The resulting dense classification can be visualized using a custom color map, enabling precise scene understanding in autonomous driving, medical imaging, environmental monitoring, and robotic perception.

Adversarial detection complements robustness defenses by flagging likely perturbed inputs. Early methods used predictive entropy and mutual information \cite{smith2018uncertainty_detection}, but adaptive attacks can bypass many detectors \cite{carlini2017adversarial_not_detected}. More advanced strategies include graph-based analysis of latent activations \cite{abusnaina2021lng_detection} and denoising comparisons \cite{grosse2017statistical_detection}. For a broader overview, see \cite{costa2023survey}.
\vspace{-0.1cm}
\textbf{General Approach.} Detection identifies if $x$ is adversarial for a classifier $F(x)$, enabling intervention in semi-autonomous systems.

\textbf{Metrics.}  
- \textit{Confidence Score:} $F(x)_{\hat{y}}$, often unreliable \cite{ma2019robust_detection}.  
- \textit{Non-Maximal Entropy (non-ME):} $\text{non-ME}(x) = \sum_{i \neq \hat{y}} F(x)_i \log(F(x)_i)$ \cite{ma2019robust_detection}.  
- \textit{Kernel Density (K-density):} $KD(x) = \sum_{x_i \in X_{\hat{y}}} k(z_i, z)$, integrating confidence and non-ME \cite{ma2019robust_detection}.

\textbf{Thresholding.} Inputs with metric above threshold $T$ are classified as normal; otherwise flagged as adversarial \cite{ma2019robust_detection}.

\textbf{Challenges and Solutions.} Adversarial detection is difficult due to attack prevalence and potential degradation of normal accuracy. Ma et al. \cite{ma2019robust_detection} combine thresholding with K-density to detect attacks effectively with minimal extra training.

\subsection{Adversarial Attack Detector}

\noindent\textbf{Pre-processing.}\hspace{0pt} Input images are randomly cropped and resized to $224 \times 224$, flipped horizontally with 50\% probability, and normalized using ImageNet’s mean and standard deviation:
\[
\mu = [0.485, 0.456, 0.406], \quad
\sigma = [0.229, 0.224, 0.225]
\]
These transformations standardize the inputs and improve model robustness.

\noindent\textbf{Model Setup.}\hspace{0pt} The Adversarial Attack Detector is trained on 100 classes using pre-trained ResNet-18 and ResNet-50. The final fully connected layer is replaced with a 102-class output, while all other layers are frozen. After training, the model is saved and later reloaded for inference with visual examples of ResNet-18 predictions shown in Figure~1. A secondary 2-class model is then created by slicing weights from the 102-class model. Finally, inference and visualization are performed on unseen images to display predictions.

\begin{figure}[H]
    \centering
    \begin{subfigure}[t]{0.6\linewidth}
        \centering
        \includegraphics[width=\textwidth,height=0.9\textwidth]{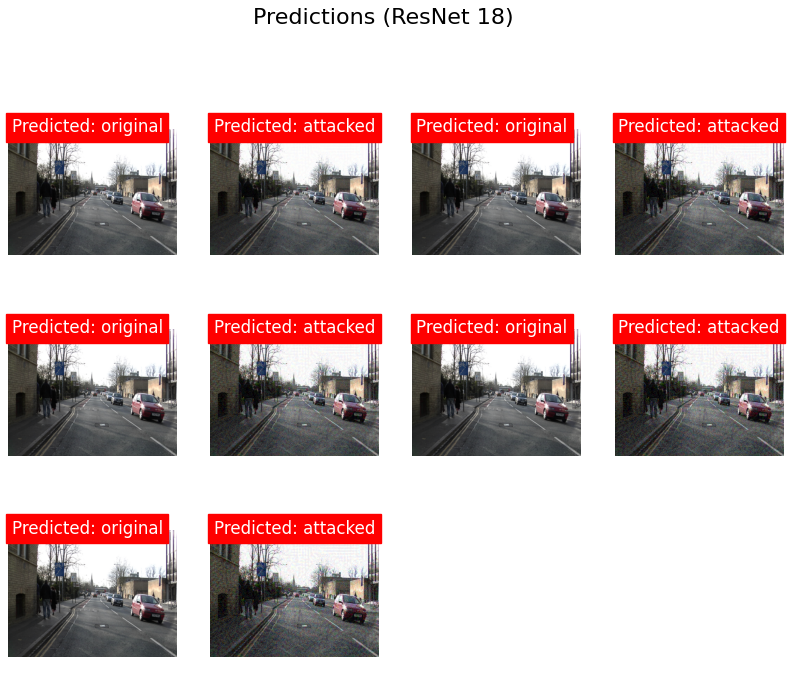}
        \caption{ResNet‑18}
        \label{fig:comparison18}
    \end{subfigure}
    \hfill
    \begin{subfigure}[t]{0.6\linewidth}
        \centering
        \includegraphics[width=\textwidth,height=0.9\textwidth]{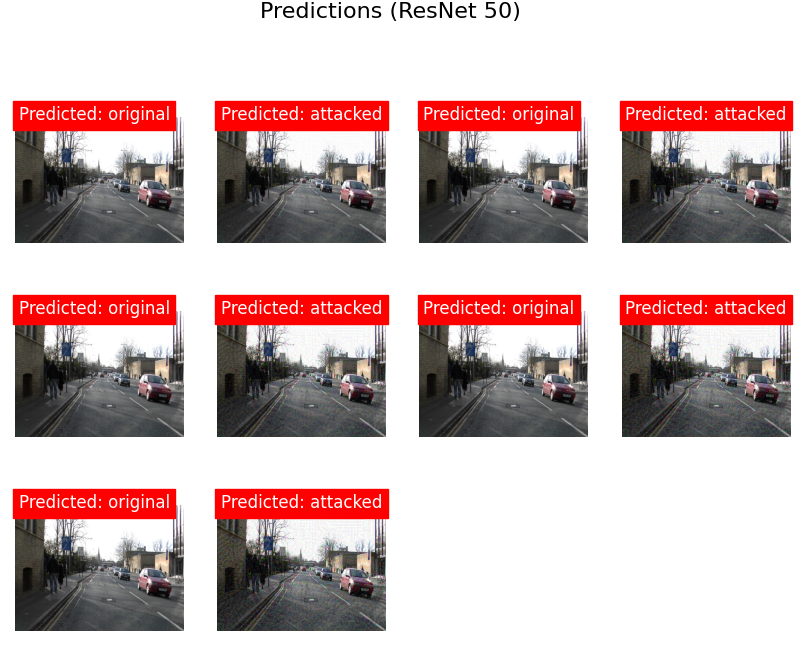}
        \caption{ResNet‑50}
        \label{fig:comparison50}
    \end{subfigure}
    \caption{Classifier predictions on clean and adversarial images. (a) ResNet‑18 shows unstable predictions with frequent misclassifications. (b) ResNet‑50 demonstrates robust, consistent performance against adversarial perturbations.}
    \label{fig:comparison}
\end{figure}

\subsection{Training}

\textbf{Loss \& Optimization.} Cross-Entropy Loss measures the discrepancy between predicted probabilities and true labels. Model parameters are updated via Stochastic Gradient Descent (SGD) with learning rate 0.001 and momentum 0.9, applied only to the classifier.

\textbf{Inference.} For unseen images, predicted class labels are obtained using the $\arg\max$ of output probabilities.

\textbf{Model Architecture.} A ResNet backbone (ResNet‑18 or ResNet‑50) with optional ImageNet-pretrained weights is used. The original fully-connected head is replaced with a linear classifier for the target number of classes (default 100), freezing all backbone parameters. Inputs are pre-processed with random resized crop ($224 \times 224$), center crop for validation, and normalized with ImageNet mean and std. Batch size is 4. For binary adversarial detection, a 2-logit head is created by copying the first two rows of the trained classifier. This performance difference is further visualized in the classifier predictions for ResNet-18 Figure~\ref{fig:comparison} and ResNet-50 Figure~\ref{fig:original_loss3}, demonstrating the models' responses to inputs

\raggedbottom

\section{Performance Evaluation}
\label{sec:performance_evaluation}

\textbf{Evaluation Metrics.} Cross-Entropy Loss ($L_{CE}$) measures the discrepancy between predicted probabilities and true labels, ignoring “ignore” labels. Intersection over Union (IoU) computes the ratio of true positives to the union of predicted and ground-truth pixels, with mean IoU (mIoU) averaging over all classes. The Dice/F1 Score accounts for true positives, false positives, and false negatives. Pixel Accuracy (PA) gives the proportion of correctly classified pixels, excluding “ignore” labels.

\textbf{Lost Classes.} Lost Classes are identified by comparing baseline class sets with adversarial predictions, highlighting classes that were misclassified or omitted during attacks, providing insight into model robustness under perturbations.

The impact of increasing FGSM strength (e) on segmentation metrics for the detector is quantified in Table~\ref{tab:tab1}. This degradation, where accuracy and mIoU decrease as e increases, is visually represented in Figure~\ref{fig:original_loss3}

The trends in ResNet-50's validation performance metrics, including accuracy, loss, and F1-score, are graphically represented in Figure~\ref{fig:resnet18_val_first10}.

ResNet-50 outperforms ResNet-18 in all metrics (gains 0.04–0.06) and shows lower loss, as its greater depth enables extraction of more complex features, reducing generalization error and improving overall accuracy.

Some individual \texttt{(epoch, phase)} rows show large gaps (up to $0.6$ in accuracy), including:

\begin{itemize}
    \item Epoch 43, \texttt{train}: ResNet-18 = 0.2 vs. ResNet-50 = 0.8 ($\Delta = -0.6$)
    \item Epoch 1, \texttt{train}: ResNet-18 = 0.2 vs. ResNet-50 = 0.8 ($\Delta = -0.6$)
    \item Epoch 29, \texttt{train}: ResNet-18 = 0.4 vs. ResNet-50 = 1.0 ($\Delta = -0.6$)
    \item Epoch 16, \texttt{train}: ResNet-18 = 0.4 vs. ResNet-50 = 1.0 ($\Delta = -0.6$)
    \item Epoch 40, \texttt{train}: ResNet-18 = 0.6 vs. ResNet-50 = 1.0 ($\Delta = -0.4$)
\end{itemize}

\begin{table}[H]
\centering
\begin{minipage}{0.48\linewidth}
\centering
\caption{FGSM Attack Metrics}
\label{tab:tab1}
\begin{tabular}{rrrrrrr}
\toprule
$\epsilon$ & pixel\_acc & mIoU & PA & mAcc & mIoU\_agg & mF1 \\
\midrule
0.00 & 1.00 & 1.00 & 1.00 & 1.00 & 1.00 & 1.00 \\
0.02 & 0.78 & 0.48 & 0.78 & 0.56 & 0.48 & 0.59 \\
0.04 & 0.65 & 0.26 & 0.65 & 0.33 & 0.26 & 0.32 \\
0.05 & 0.61 & 0.19 & 0.61 & 0.29 & 0.19 & 0.24 \\
0.06 & 0.59 & 0.19 & 0.59 & 0.26 & 0.19 & 0.25 \\
0.07 & 0.57 & 0.17 & 0.57 & 0.23 & 0.17 & 0.22 \\
0.08 & 0.54 & 0.13 & 0.54 & 0.19 & 0.13 & 0.18 \\
0.09 & 0.52 & 0.11 & 0.52 & 0.18 & 0.11 & 0.15 \\
0.10 & 0.49 & 0.10 & 0.49 & 0.16 & 0.10 & 0.13 \\
\bottomrule
\end{tabular}
\end{minipage}
\hfill
\begin{minipage}{0.48\linewidth}
\centering
\includegraphics[width=1.2\linewidth]{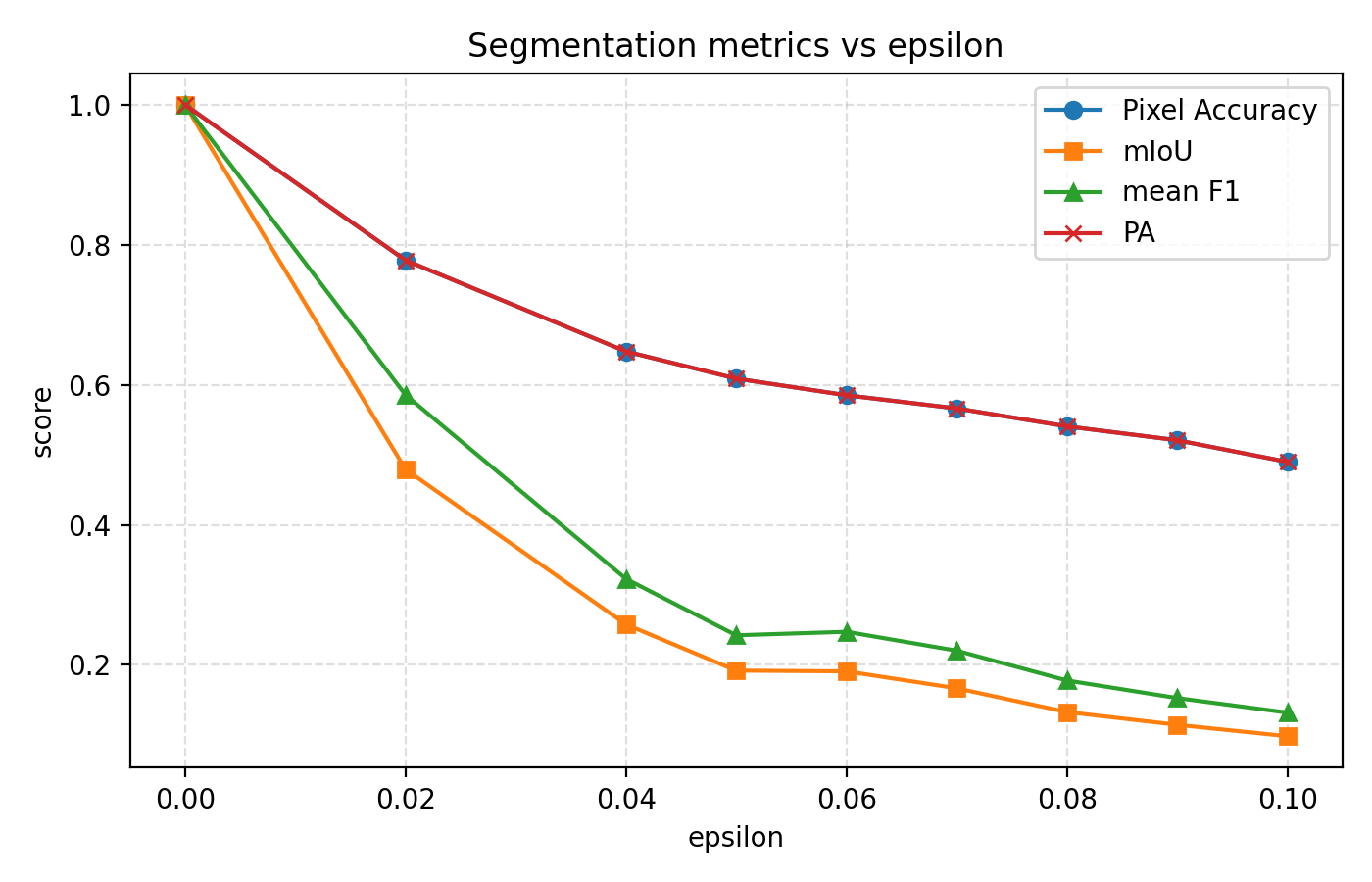}
\captionof{figure}{Effect of increasing FGSM attack strength ($\epsilon$) on segmentation performance. As $\epsilon$ rises, accuracy and mIoU drop, and several classes are completely lost.}
\label{fig:original_loss3}
\end{minipage}
\end{table}

Table~\ref{tab:tab1} and Figure~\ref{fig:original_loss3} show that FGSM attacks sharply reduce performance: mIoU drops from 1.00 to 0.48 at e = 0.02 and to 0.10 at e = 0.10, highlighting the need for the proposed detection framework.

\begin{table}[H]
\centering
\begin{minipage}{0.48\linewidth}
\centering
\caption{ResNet-18 Validation Metrics (Epochs 1--10)}
\label{tab:table2}
\begin{tabular}{@{}ccccccc@{}}
\toprule
Epoch & Phase & Loss & Accuracy & Precision & Recall & F1-Score \\
& & & & (Macro) & (Macro) & (Macro) \\
\midrule
1  & val & 0.5 & 1.0 & 1.0 & 1.0 & 1.0 \\
2  & val & 0.6 & 0.7 & 0.7 & 0.5 & 0.8 \\
3  & val & 0.5 & 0.7 & 0.7 & 0.5 & 0.8 \\
4  & val & 0.9 & 0.7 & 0.7 & 0.5 & 0.8 \\
5  & val & 1.1 & 0.7 & 0.7 & 0.5 & 0.8 \\
6  & val & 1.0 & 0.7 & 0.7 & 0.5 & 0.8 \\
7  & val & 0.5 & 0.7 & 0.7 & 0.5 & 0.8 \\
8  & val & 0.4 & 0.7 & 0.7 & 0.5 & 0.8 \\
9  & val & 0.5 & 1.0 & 1.0 & 1.0 & 1.0 \\
10 & val & 0.3 & 1.0 & 1.0 & 1.0 & 1.0 \\
\bottomrule
\end{tabular}
\end{minipage}
\hfill
\begin{minipage}{0.44\linewidth}
\centering
\includegraphics[width=1.4\linewidth]{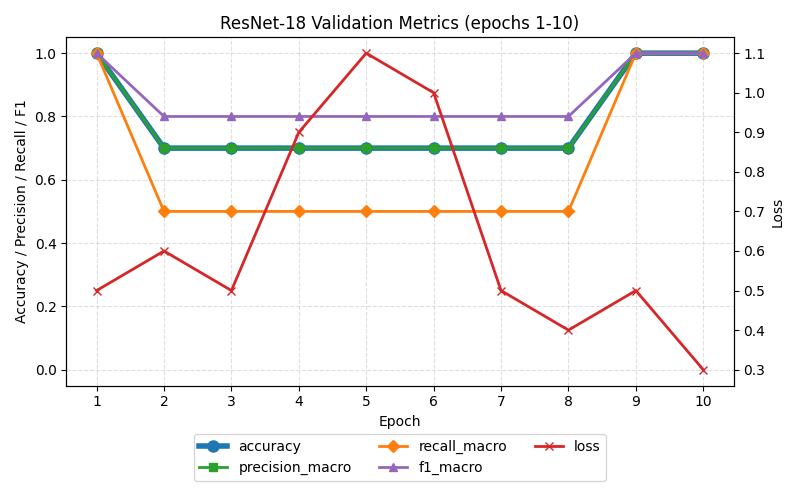}
\captionof{figure}{ResNet-18 validation performance across epochs 1--10, including accuracy, precision, recall, F1-score, and loss.}
\label{fig:resnet18_val_first10}
\end{minipage}
\end{table}

The validation metrics for the ResNet-18 model across the initial ten epochs are detailed in Table~\ref{tab:table2}

These validation results are also plotted for visualization, as shown in Figure~\ref{fig:resnet18_val_first10}.

In contrast, the validation performance for the ResNet-50 model during epochs 1-10 is provided in Table~\ref{tab:tab3}

\begin{table}[H]
\centering
\begin{minipage}{0.48\linewidth}
\centering
\caption{ResNet-50 Validation Metrics (Epochs 1--10)}
\label{tab:tab3}
\begin{tabular}{@{}ccccccc@{}}
\toprule
Epoch & Phase & Loss & Accuracy & Precision & Recall & F1-Score \\
& & & & (Macro) & (Macro) & (Macro) \\
\midrule
1  & val & 0.7 & 0.7 & 0.7 & 0.5 & 0.8 \\
2  & val & 0.8 & 0.7 & 0.7 & 0.5 & 0.8 \\
3  & val & 1.2 & 0.7 & 0.7 & 0.5 & 0.8 \\
4  & val & 1.3 & 0.7 & 0.7 & 0.5 & 0.8 \\
5  & val & 0.9 & 0.7 & 0.7 & 0.5 & 0.8 \\
6  & val & 0.6 & 0.7 & 0.7 & 0.5 & 0.8 \\
7  & val & 0.7 & 0.3 & 0.3 & 0.5 & 0.5 \\
8  & val & 0.6 & 0.7 & 0.8 & 0.8 & 0.7 \\
9  & val & 0.5 & 0.7 & 0.7 & 0.5 & 0.8 \\
10 & val & 0.7 & 0.7 & 0.7 & 0.5 & 0.8 \\
\bottomrule
\end{tabular}
\end{minipage}
\hfill
\begin{minipage}{0.45\linewidth}
\centering
\includegraphics[width=1.2\linewidth]{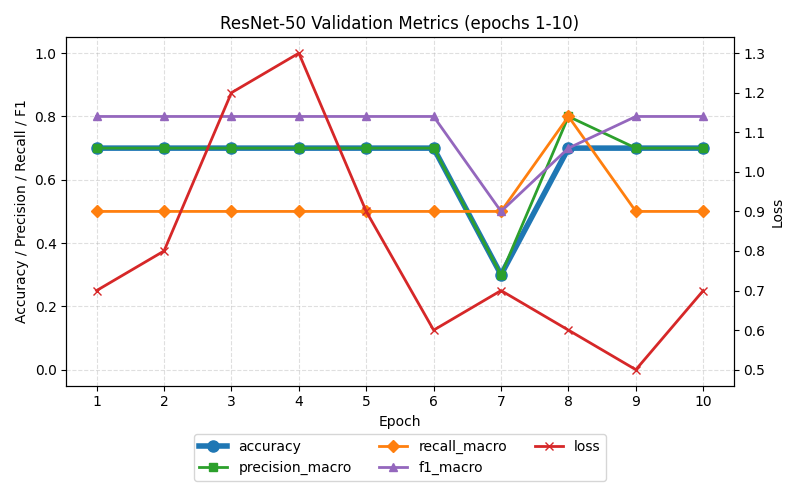}
\captionof{figure}{ResNet-50 validation performance across epochs 1--10, including accuracy, precision, recall, F1-score, and loss.}
\label{fig:resnet50_val_first10}
\end{minipage}
\end{table}

Large accuracy swings (up to 0.6) reveal training instability in ResNet-18, with brief peaks followed by drops, indicating convergence issues. ResNet-50, in contrast, shows stable, consistent learning, offering greater robustness and better resilience to adversarial perturbations for robotic perception tasks.

\begin{figure}[H]
    \centering
    \vspace{0pt}
        \centering
        \includegraphics[width=1\linewidth,height=0.6\linewidth]{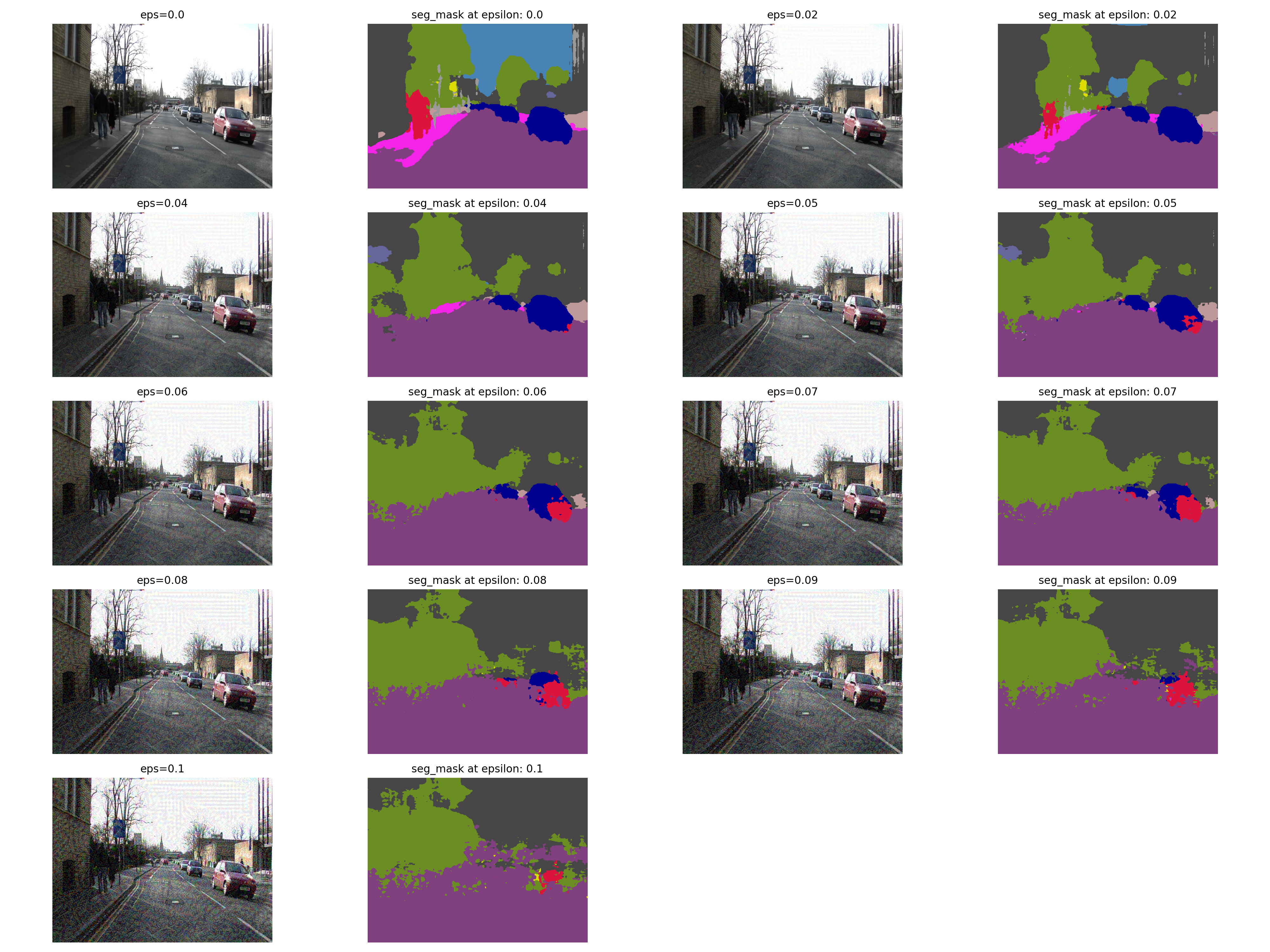}
        \caption{Visualization of FGSM adversarial examples \cite{goodfellow2015explaining} on a DeepLabV3+ \cite{chen2018encoder} model with ResNet-18 \cite{he2016deep} \cite{deng2009imagenet}  \cite{cordts2016cityscapes}, showing clean and adversarial inputs with increasing $\epsilon$, \cite{paszke2019pytorch}.}
        \label{fig:original_loss2}
\end{figure}

Further analysis includes the visualization of FGSM adversarial examples, illustrating clean and perturbed inputs across increasing e values (Figure~\ref{fig:original_loss2}).

Architectural differences affect adversarial detection. ResNet-50 offers better baseline segmentation but may be more vulnerable, requiring a stricter detection threshold than ResNet-18.

Figure~\ref{fig:original_loss2} shows that small increases in attack strength e quickly degrade segmentation, with mIoU dropping from 1.00 to 0.10 (Table 1). Despite minimal visual changes, this highlights the need for detection based on deep feature analysis rather than visual inspection.

\section{Conclusions}
\label{sec:conclusions}

This paper addresses the challenge of detecting adversarial attacks on robotic perception systems, focusing on semantic pixel-wise segmentation with pre-trained ResNet-18 and ResNet-50 models. We demonstrate a method that distinguishes original from adversarially manipulated images, showing through extensive experiments that it enhances segmentation model robustness.

The results highlight the importance of incorporating security into automated systems and provide a foundation for improving resilience against adaptive adversarial attacks. As AI expands in safety-critical applications, ensuring reliability and trustworthiness is essential. By advancing adversarial detection and countermeasures, this work aims to build confidence in intelligent systems and maximize their societal benefits.

In conclusion, the study lays groundwork for secure robotic perception systems and emphasizes the need for ongoing research to improve model performance while safeguarding against adversarial vulnerabilities for safe real-world deployment.

\bibliographystyle{splncs04}  
\bibliography{references}

\end{document}